\documentclass[letterpaper, 10 pt, conference]{ieeeconf}  

\IEEEoverridecommandlockouts                              

\usepackage{amsmath}
\usepackage{amssymb}
\usepackage{mathtools}
\usepackage{graphicx}
\usepackage{float}
\usepackage{subcaption}
\usepackage{epstopdf}
\usepackage[tableposition=top]{caption}
\usepackage{array}
\usepackage{url}
\usepackage{algorithm}
\usepackage{algpseudocode}
\usepackage{diagbox}
\usepackage[table]{xcolor}
\usepackage[framemethod=TikZ]{mdframed}
\usepackage[procnames]{listings}
\usepackage[hidelinks]{hyperref}

\mdfdefinestyle{frameStyle}{%
    roundcorner=5pt,
    backgroundcolor=white}

\algtext*{EndIf}
\algtext*{EndFor}
\algtext*{EndWhile}
\algtext*{EndFunction}

\newcommand{\assign}[0]{$\leftarrow$ }
\newcommand{\algto}[0]{\textbf{to} }
\newcommand{\algin}[0]{\textbf{in} }

\title{\LARGE \bf
Robot Action Diagnosis and Experience Correction\\by Falsifying Parameterised Execution Models
}

\author{Alex Mitrevski$^{\dagger\mathsection}$, Paul G. Pl{\"o}ger$^{\dagger}$, and Gerhard Lakemeyer$^{\ddagger}$
\thanks{$^{*}$This work was supported by the b-it Foundation.} %
\thanks{$^{\dagger}$Alex Mitrevski and Paul G. Pl{\"o}ger are with the Department of Computer Science, Hochschule Bonn-Rhein-Sieg, Sankt Augustin, Germany\newline
        {\tt\scriptsize <aleksandar.mitrevski, paul.ploeger>@h-brs.de}} %
\thanks{$^{\ddagger}$Gerhard Lakemeyer is with the Department of Computer Science, RWTH Aachen University, Aachen, Germany
        {\tt\scriptsize gerhard@informatik.rwth-aachen.de}} %
\thanks{$^{\mathsection}$Corresponding author} %
}

\begin{document}

\maketitle
\thispagestyle{empty}
\pagestyle{empty}


\begin{abstract}

    When faced with an execution failure, an intelligent robot should be able to identify the likely reasons for the failure and adapt its execution policy accordingly. This paper addresses the question of how to utilise knowledge about the execution process, expressed in terms of learned constraints, in order to direct the diagnosis and experience acquisition process. In particular, we present two methods for creating a synergy between failure diagnosis and execution model learning. We first propose a method for diagnosing execution failures of parameterised action execution models, which searches for action parameters that violate a learned precondition model. We then develop a strategy that uses the results of the diagnosis process for generating synthetic data that are more likely to lead to successful execution, thereby increasing the set of available experiences to learn from. The diagnosis and experience correction methods are evaluated for the problem of handle grasping, such that we experimentally demonstrate the effectiveness of the diagnosis algorithm and show that corrected failed experiences can contribute towards improving the execution success of a robot.

\end{abstract}


\section{INTRODUCTION}

    \subsection{Motivation}

    For autonomous robots, the process of acting in the real world is inevitably associated with execution failures: a robot may fail to grasp an object, spill a liquid while pouring it into a glass, or miss a keyhole while inserting a key.\footnote{Such failures often occur due to perceptual or navigation inaccuracies, but may also be the result of incomplete knowledge about performed actions.} Even when they fail, robots are generally following a parameterised policy that, according to their available knowledge, should lead to execution success. A characteristic of intelligent agents, then, should be their ability to analyse such failures more closely so that they can be used as a learning opportunity that would lead to a richer execution policy.

    There are different ways of diagnosing failures \cite{khalastchi2018}, for example using data-driven methods \cite{obry2018}, which are useful when significant data about the problem of interest are available, or common discrete model-based approaches \cite{peischl2003,dekleer1987}, which may be supplemented by sequential diagnosis \cite{rodler2018,altan2014} or preference relations \cite{bouziat2018}. In typical robot applications, the requirements for applying such methods do not always hold, namely a diagnosis model or sufficient data for learning one may be unavailable.\footnote{Additionally, most learning paradigms can only identify data associations rather than causal relations, which are required for meaningful diagnosis.} A related way of finding diagnoses is using counterfactual reasoning \cite{pearl2011}, where alternative aspects of the world and their consequences are explored, and which has been shown to improve the quality of diagnoses compared to conventional learning methods \cite{richens2020}.

\begin{figure}[tp]
        \centering
        \vspace{0.22cm}
        \includegraphics[width=\linewidth]{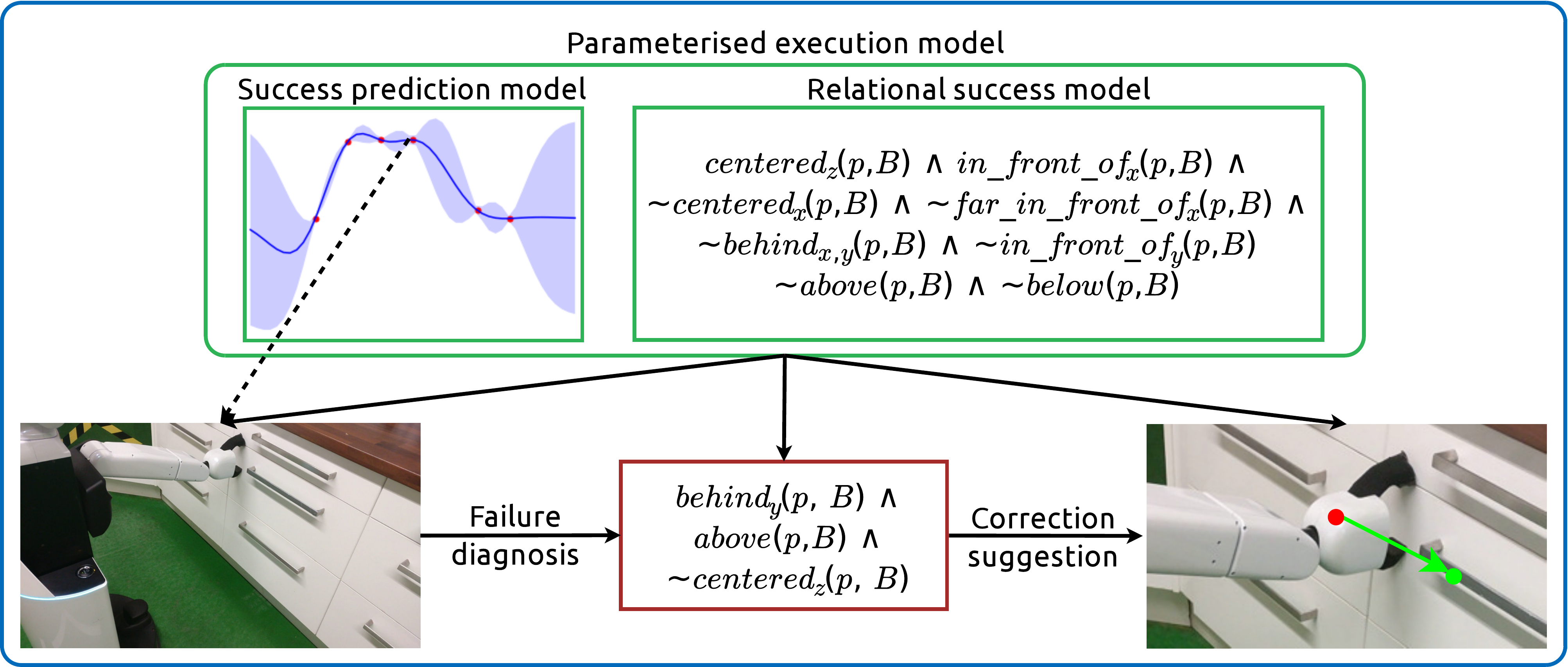}
        \caption{Diagnosis and experience correction framework proposed in this paper. The relational success model shown here is an enhanced version of the model learned in \cite{mitrevski2020}; the diagnosis represents a violation of the model.}
        \label{fig:framework}
    \end{figure}

    In this paper, we apply execution models of parameterised actions as in \cite{mitrevski2020} and investigate their use for failure diagnosis and experience correction through techniques inspired by counterfactual reasoning \cite{pearl2011} and qualitative modelling \cite{struss1996}. We introduce methods for (i) diagnosing failures by perturbing the parameters of a failed action to identify relations that the parameterisation is close to violating and (ii) proposing failure corrections by moving the parameterisation away from the region where the violation of the preconditions was identified. An overview of the proposed framework is given in Fig. \ref{fig:framework}. We evaluate our algorithms on the use case of grasping furniture handles with a Toyota Human Support Robot (HSR) \cite{hsr_paper}. On a set of failed handle grasps, we first investigate how different parameters of the diagnosis algorithm affect the accuracy of the identified diagnoses. We then demonstrate that a success prediction model learned purely from experiences corrected by our proposed algorithm can lead to a reasonable execution success rate of a robot, which supports our hypothesis that understanding failed executions is important for intelligent robot execution policies.\footnote{Accompanying video: \url{https://youtu.be/xroO0xvhpdo}}

    \subsection{Background}
    \label{sec:background}

    In our prior work \cite{mitrevski2020}, we introduced a representation of execution models of parameterised actions, which can be seen as a hybrid representation of an execution policy. Formally, an execution model $M$ comprises a model $R = (R_1, ..., R_m)$ of relational constraints on the action parameters, potentially under a set of $m$ qualitative modes, and a continuous model $F$ that encodes the predicted success likelihood for an action parameterisation $\mathbf{x}$, optionally under constraints on the action outcomes. Fig. \ref{fig:framework} illustrates an execution model for handle grasping. An execution model is learned from experience, such that $R$ is extracted from a predefined set of relations for each action, while $F$ is represented by a Gaussian process. For execution, parameters are sampled from $F$ using rejection sampling until a sample that satisfies the relational model $R$ is found. In this paper, we make use of this representation and show how it can be utilised for diagnosing and correcting execution failures.


\section{RELATED WORK}
\label{sec:related_work}

    Failure diagnosis and subsequent learning from failures are rarely considered in conjunction; nevertheless, our work draws upon various aspects from the literature addressing these two problems.

    Hermann et al. \cite{hermann2020} describe a simulation-based curriculum learning method, which learns an initial policy from human demonstrations and then changes the complexity of the learned task by starting near the goals of the demonstrations and successively increasing the distance from the goal as the robot becomes better at the task. Wang and Kroemer \cite{wang2019} propose a strategy for improving demonstrated skills by using information about contacts with the environment; in particular, skill experiences are collected by executing the task as originally demonstrated, but by perturbing the end effector's position at the end of each skill, which results in a set of experiences that additionally include contact modes. Our strategy for correcting failed action parameterisations uses a somewhat similar idea, as it modifies the action parameters around the point of interest, but we explicitly constrain updates by the known action preconditions.

    In the context of learning trajectories from demonstration, Grollman and Billard \cite{grollman2012} consider the problem of using failed demonstrations during learning. For this purpose, a Donut distribution is defined and used, which allows representing state space holes, namely regions of the space that should be avoided when sampling trajectories. Haidu et al. \cite{haidu2015} consider learning a trajectory model from multiple demonstrations that can be used for verifying whether a trajectory that a robot is trying to execute matches the model, which is represented as a trajectory envelope. Mueller et al. \cite{mueller2018} present a learning by demonstration framework that allows combining potentially suboptimal/faulty trajectory demonstrations with successful demonstrations that also encode task constraints, which are represented by logical predicates about certain task-relevant aspects. As in \cite{grollman2012,mueller2018}, our objective is to explicitly incorporate information about failures into the success model. Similarly, as in \cite{haidu2015,mueller2018}, we make use of a constraint model, but while the model there is only used to prevent failures, we additionally identify corrective action parameterisations when failures do occur.

    In the context of diagnosis, Parker and Kannan \cite{parker2006} present a diagnosis algorithm that uses a manually specified causal model that relates faults in a multi-robot system. Given such a model, fault symptoms are created by monitoring individual components; diagnosis is then performed using case-based reasoning and active testing, followed by a recovery procedure and potentially an update of the causal model for including new faults. Zaman and Steinbauer \cite{zaman2013} propose a method for creating observers and extracting a diagnosis model from failure-free data from a robot that uses a component-based, communication-oriented software architecture. In particular, communication patterns between system components are used for learning a nominal model of execution; runtime violations of the model then trigger a model-based diagnosis procedure. Inspired by \cite{parker2006}, our diagnosis and experience correction methods use a weak causal model that relates action parameters to relations that they affect. As in \cite{zaman2013}, we use a learned nominal model of execution for diagnosis, but the failures we are interested in cannot be identified by analysing communication data.

    Our work also has a conceptual relation to safe reinforcement learning \cite{garcia2015}, but while the objective there is to avoid failures as much as possible, we hypothesise that at least some failures are inevitable when acting in human-centered environments, so a robot should be able to use those for informing its learning and execution processes.


\section{FAILURE DIAGNOSIS AND CORRECTION}
\label{sec:diagnosis}

    In this section, we describe the idea behind the proposed diagnosis and experience correction methods. We also briefly discuss how the nature of the known preconditions affects the failures that can be diagnosed with our method.

    \begin{figure*}[tp]
        \begin{subfigure}{0.475\linewidth}
            \centering
            \includegraphics[width=\linewidth]{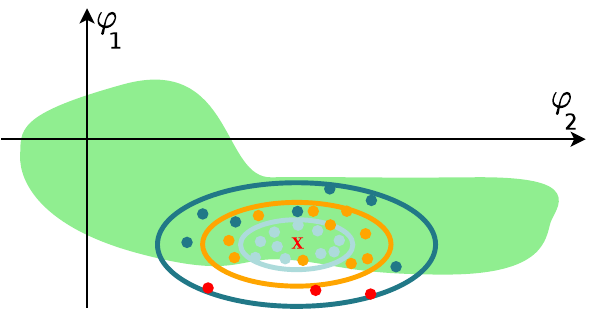}
            \caption{An illustration of the search process for action precondition violations. The concentric ellipses represent increasingly larger violation search regions, and the points of the same colour as the ellipses represent samples within a search region. The search process is performed until non-conflicting violations are found within a search region.}
            \label{fig:relation_violation_search}
        \end{subfigure}
        \hspace{0.025\textwidth}
        \begin{subfigure}{0.475\linewidth}
            \centering
            \includegraphics[width=\linewidth]{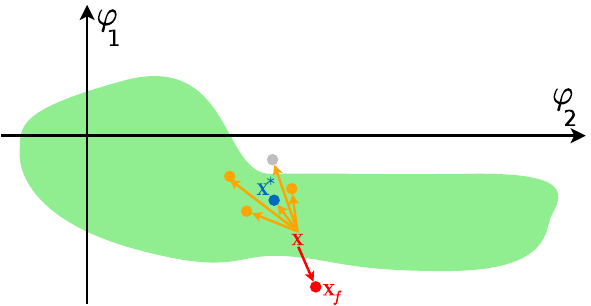}
            \caption{An illustration of the correction of execution parameters. The orange arrows are parameterisations that face away from the red point. The gray point violates the preconditions and is not a valid correction candidate. The point that is selected as a corrective experience, illustrated here in blue, is the one maximising the predicted execution success under the known execution model.}
            \label{fig:experience_correction}
        \end{subfigure}
        \caption{Diagnosing and correcting an execution failure. $\varphi_1$ and $\varphi_2$ are two action parameters and $\mathbf{x}$ marks parameters that have caused a failure. The green region marks the parameter space where, according to the robot's model, the preconditions are satisfied; the red points fall outside and thus represent parameterisations that violate at least one of the preconditions.}
    \end{figure*}

    \subsection{Notation}

    We denote a zero vector of length $m$ by $\mathbf{0}^{m}$ and a zero matrix of size $m \times n$ by $0^{m \times n}$. $\mathbf{x}_i$ is the $i$-th entry of vector $\mathbf{x}$. $A_{i}$ is the $i$-th row and $A_{ij}$ the $(i, j)$-th entry of matrix $A$.

    \subsection{Diagnosing Failed Executions}

    We propose a failure diagnosis method that, given an action parameterisation that has lead to a failure, searches for preconditions that are likely to have been violated, or are close to being violated, by the parameterisation. The intuition behind the method is illustrated in Fig. \ref{fig:relation_violation_search}, and a detailed description of the working principles is given below.

    Let us denote the failed action parameterisation by $\mathbf{x}$. The diagnosis algorithm samples $k_{\max}$ alternative parameterisations $\mathbf{x}'_k, 1 \leq k \leq k_{\max}$ around $\mathbf{x}$, such that $\mathbf{x}'_k \sim \mathcal{N}(\mathbf{x}, \Sigma)$. For sampling from the Gaussian distribution, we take a diagonal covariance matrix $\Sigma$ with a predefined initial set of standard deviations for each parameter $\varphi_i$ of $\mathbf{x}$, $1 \leq i \leq |\mathbf{x}|$. The result is (i) a set $D$ of relations that are likely diagnosis candidates, as well as (ii) a parameterisation $\mathbf{x}_f$ that satisfies $D$ and thus violates the action preconditions. To find these, the algorithm maintains (i) a set $D$ of relations that, under $\mathbf{x}'$, violate the precondition model $R_q$ for a given qualitative mode $q$ and are thus potential diagnosis candidates, and (ii) a set $X$ of $\mathbf{x}'$ that associates each action parameter $\varphi \in \Phi$ with the parameterisation that caused the violation.

    When associating a violated precondition $p$ with a parameterisation $\mathbf{x}'$, it should be noted that different parameters have an effect on different relations.\footnote{For instance, when grasping an object, moving the end effector to the left/right only affects relations such as \texttt{leftOf}/\texttt{rightOf} and has no effect on relations such as \texttt{above} or \texttt{inFrontOf}.} Because of this, the complete parameterisation $\mathbf{x}'$ is not of interest when looking for violations of the preconditions, but only the values of the parameters that actually affect the violated relations. We particularly assume that each relation is only affected by a single action parameter $\varphi$; this means that we have a surjective mapping $\mathcal{M}: P \rightarrow \Phi$ that maps relations to parameters that affect them.\footnote{The mapping of parameters to relations can, in principle, be learned from experience, but this aspect is not in the scope of this paper.} Given $\mathcal{M}$ and a violated precondition $p$, $X$ is updated with a vector $\mathbf{x}_{0}'$, which has zeros everywhere except at the position of the parameter $\varphi$ that affects $p$. This allows us to create the resulting parameterisation $\mathbf{x}_f$ that affects all relations in $D$ by a simple summation of the elements in $X$:
    \begin{equation}
        \mathbf{x}_f = \sum_{k}X_{k}
        \label{eq:parameterCombination}
    \end{equation}

    Depending on the size of the search region, different samples $\mathbf{x}'$ may lead to potentially contradicting relation violations.\footnote{For instance, in the case of grasping the narrow handle in Fig. \ref{fig:framework}, both \texttt{above} and \texttt{below} should be false for successful execution of the action, but if the search region is larger than the size of the handle, both violations may be found within the $k_{\max}$ samples, even though they obviously cannot hold at the same time. To be able to find such contradictions between relations, we explicitly annotate disjoint relations in the action model.} To prevent this case, $D$ is post-processed each time a precondition violation is found, such that any contradicting relations are removed from the set. To maintain the consistency of $X$, the parameterisations that were responsible for adding those relations to $D$ are also removed from $X$.

    If no diagnosis candidates are found after generating all $k_{\max}$ samples, the search region is expanded by a predefined ratio $r$, which amounts to increasing the magnitude of each value in $\Sigma$ for the $i+1$-th iteration of the algorithm:
    \begin{equation}
        \Sigma_{i+1} = \Sigma_{i} + r\Sigma_{i}
    \end{equation}

    Algorithm \ref{alg:failedActionDiagnosis} summarises the method. Conceptually, the idea behind this algorithm is similar to that of iterative deepening tree search \cite{korf1987}, but applied to a continuous parameter space.
    \begin{algorithm}[tp]
        \begin{algorithmic}[1]
            \Function{\texttt{diagnoseFailure}}{$R$, $q$, $\mathbf{x}$, $k_{\max}$, $\Sigma$, $r$}
                \State $D$ \assign $\varnothing$
                \While{$D = \varnothing$}
                    \State $X$ \assign $\varnothing$
                    \For{$k$ \assign $1$ \algto $k_{\max}$}
                        \State $\mathbf{x}'$ \assign $\mathcal{N}(\mathbf{x}, \Sigma)$
                        \State $\mathbf{p}'$ \assign \texttt{extractRelations}$(\mathbf{x}')$
                        \If{$\mathbf{p}' \neq R_q$}
                            \State $C$ \assign $\{ p | p \in \mathbf{p}', p \not\in R_q \}$
                            \State $D$ \assign $D \cup C$
                            \State $X$ \assign \texttt{updateParameters}($X$, $C$, $\mathbf{x}'$)
                            \State ($D$, $X$) \assign \texttt{removeConflicts}($D$, $X$)
                        \EndIf
                    \EndFor
                    \If{$D = \varnothing$}
                        \State $\Sigma$ \assign $\Sigma + r\Sigma$
                    \EndIf
                \EndWhile
                \State $\mathbf{x}_f$ \assign $\sum_{k}X_{k}$
                \State \Return ($D$, $\mathbf{x}_f$)
            \EndFunction
        \end{algorithmic}
        \caption{Diagnosing a failed parameterised action by looking for alternative parameterisations that falsify the preconditions. \texttt{extractRelations} converts $\mathbf{x}$ to a relational form. \texttt{updateParameters} uses $\mathcal{M}$ to update $X$ with $\mathbf{x}_{0}'$. \texttt{removeConflicts} ensures that the diagnosis candidate set does not have contradictory relations.}
        \label{alg:failedActionDiagnosis}
    \end{algorithm}
    It should be noted that, as described above, the diagnosis algorithm terminates only when it finds at least one violation of the execution constraints; depending on the desired certainty of the diagnoses, this behaviour could be changed by adding an upper bound on the number of times the search region is updated. Due to the sampling nature of the diagnosis method, the results may differ slightly between runs. To ensure stability of the diagnoses, in practice we run the algorithm $n$ times; the relations in $D$ are then the ones whose proportion over the runs is larger than a threshold $\alpha$. The falsifying parameterisation $\mathbf{x}_f$ is similarly taken as the average of the updates of the relations in $D$ over the $n$ runs.

    \subsection{Experience Correction}
    \label{sec:diagnosis_experience_correction}

    Given the set of diagnosis candidates, explanations for a failure can be generated, but it is also important that a robot corrects its failure and, ideally, learns from it. In this section, we describe a method that uses $D$ and $\mathbf{x}_f$ for exploring alternative parameterisations $\mathbf{x}'$ that move the execution sample further inside the region where the preconditions are satisfied. The general idea behind the method is illustrated in Fig. \ref{fig:experience_correction} and a description is provided below.

    For correcting a failed execution sample, the objective is to move the parameters $\mathbf{x}$ in a direction opposite of $\mathbf{x}_f$ since, intuitively, parameters away from $\mathbf{x}_f$ are more likely to satisfy the execution preconditions; this, in turn, produces a set of synthetic learning experiences. As for diagnosis, we use a sampling algorithm for experience correction, namely we generate $s_{\max}$ alternative parameterisations $\mathbf{x}'_j, 1 \leq j \leq s_{\max}$ away from $\mathbf{x}_f$ and, for each of these, calculate the predicted success likelihood, which we will denote $s_j$, using the success model $F$. The corrected experience $\mathbf{x}^{*}$ is then the one that maximises the success likelihood, namely
    \begin{equation}
        \mathbf{x}^{*} = \mathbf{x}'_{\underset{1 \leq j \leq s_{\max}}{\arg\max}s_j}
    \end{equation}

    For generating $\mathbf{x}'_j$, we first find the direction in which $\mathbf{x}$ was perturbed to obtain $\mathbf{x}_f$; this is given as $\Delta\mathbf{x} = \mathbf{x}_f - \mathbf{x}$. For each $\varphi$ that was perturbed to find $\mathbf{x}_f$, we then have $\mathbf{x}_{j_\varphi}' = \mathbf{x}_{\varphi} + \mathcal{Q}_{j\varphi}$, where a parameter correction $\mathcal{Q}_{j\varphi}$ is sampled under the constraint that the likelihood of sampled values should be high around $-\Delta\mathbf{x}_{\varphi}$, which is in the exact opposite direction of $\Delta\mathbf{x}_{\varphi}$, and should smoothly decrease for larger and smaller values. In particular, corrections $\mathcal{Q}_{j\varphi}$ are sampled from a gamma distribution \cite{riley2006} $\Gamma_{\kappa, \theta}$ with density
    \begin{equation}
        P_{\kappa,\theta}(\varphi) = \frac{\varphi^{\kappa-1}e^{\frac{-\varphi}{\theta}}}{\theta^{\kappa}\Gamma(\kappa)}
    \end{equation}
    where $\Gamma(\kappa)$ is the gamma function, $\kappa$ controls the shape of the distribution, and $\theta > 0$ is the scale. The parameters of $\Gamma_{\kappa, \theta}$ provide enough flexibility for enforcing the above two constraints: for this, $\theta$ is set to $|\Delta\mathbf{x}_{\varphi}|$, $\kappa$ is set to a value $\kappa \geq 1$, and $\mathcal{Q}_j = -\texttt{sgn}(\Delta\mathbf{x}_j) \times \Gamma_{\kappa, \theta}$. The experience correction procedure is summarised in Algorithm \ref{alg:experienceCorrection}.
    \begin{algorithm}[tp]
        \begin{algorithmic}[1]
            \Function{\texttt{correctExperience}}{$(R,F)$, $q$, $\mathbf{x}$, $k_{\max}$, $\Sigma$, $r$, $n$, $\alpha$, $s_{\max}$, $\kappa$}
                \State $(D, \mathbf{x}_{f})$ \assign \texttt{diagnose}$(R, q, \mathbf{x}, k_{\max}, \Sigma, r, n, \alpha)$
                \State $\mathcal{Q}$ \assign $0^{s_{\max} \times |\mathbf{x}|}$
                \State $\mathbf{s}$ \assign $\mathbf{0}^{s_{\max}}$
                \State $\Delta\mathbf{x}$ \assign $\mathbf{x}_f - \mathbf{x}$
                \For{$d$ \algin $D$}
                    \State $\varphi$ \assign $\mathcal{M}(d)$
                    \For{$j$ \assign $1$ \algto $s_{\max}$}
                        \State $\mathcal{Q}_{j\varphi}$ \assign -\texttt{sgn}($\Delta\mathbf{x}_j$) $\times$ $\Gamma_{\kappa, \Delta\mathbf{x}_j}$
                    \EndFor
                \EndFor
                \For{$j$ \assign $1$ \algto $s_{\max}$}
                    \State $\mathbf{x}'$ \assign $\mathbf{x} + \mathcal{Q}_j$
                    \If{$\texttt{f}(R, \mathbf{x}', q)$}
                        \State $\mathbf{s}_j$ \assign $F(\mathbf{x}')$
                    \EndIf
                \EndFor
                \State \Return $\mathbf{x} + \mathcal{Q}_{\underset{1 \leq j \leq s_{\max}}{\arg\max} \; \mathbf{s}_j}$
            \EndFunction
        \end{algorithmic}
        \caption{Correcting a failed execution. \texttt{diagnose} runs the \texttt{diagnoseFailure} function given in Algorithm \ref{alg:failedActionDiagnosis} $n$ times with an acceptance threshold $\alpha$. \texttt{f} verifies that the relations $R$ are satisfied by $\mathbf{x}'$ under a qualitative mode $q$.}
        \label{alg:experienceCorrection}
    \end{algorithm}

    At this point, it is worth considering which failures can be diagnosed by our method, as this also affects the quality of the proposed experience corrections. Intuitively, the identification of failure causes depends on the relational model $R$, so the relations in $R$ and, more generally, the relations from which $R$ is extracted, have a direct effect on the diagnosability of failures.\footnote{In other words, the robot's conceptual understanding of its actions is directly proportional to its ability to diagnose and correct its own failures.} As discussed before, our method expects relations to satisfy the assumption about the surjectivity of $\mathcal{M}$. In addition, as it is difficult to guarantee that $R$ includes all necessary relations for generating meaningful diagnoses, the proposed algorithms should ideally be embedded in a continual model verification and enhancement process. In particular, the set of relations has to be expanded with additional relations by a human teacher until most failures of interest can be diagnosed, which is what we have done in this paper, or the robot itself could initiate an automatic learning method to identify relations that govern an action.\footnote{For the handle grasping example in Fig. \ref{fig:framework}, the failure in which the robot attempts to grasp at a position that is too far from the handle - which can happen due to an incorrect handle pose estimate - is not diagnosable by our algorithm if the model does not include the constraint that the handle should be within the end effector's reach. Including $far\_in\_front\_of_{x}$ in the set of relations, which allows encoding this, is an aspect in which the relational model used in this paper differs from the model presented in \cite{mitrevski2020}.}


\section{EXPERIMENTS}
\label{sec:experiments}

    We performed two experiments to investigate the effectiveness of the diagnosis and correction methods, both in the context of grasping a drawer handle with a Toyota HSR as shown in Fig. \ref{fig:framework}. Before each trial, the robot is manually positioned to face the handle. As in \cite{mitrevski2020}, the handle grasping action is parameterised by the end effector's position with respect to the center of the handle's bounding box.

    In the first experiment, the objective is to evaluate the correctness of the failure diagnosis method; for this, the causes of failed executions are manually labelled and the ground-truth labels are then compared with the results of our method. For execution, parameters are sampled randomly within the bounding box of the detected handle, except for the front direction ($x$ in terms of the robot's base frame), where values are sampled between $5cm$ and $15cm$ in front of the bounding box so that the data set also includes failures in which the robot is too far to grasp the handle.\footnote{We sample randomly rather than from a learned model because we want to have a large and diverse set of failures. For instance, some of the samples in which the grasp was performed too far from the handle violate the $far\_in\_front\_of_{x}$ precondition and would not be accepted if parameters were sampled from a learned model. They are nevertheless kept in the evaluation so that we have a richer set of failures.} We performed $100$ repetitions of the action, out of which the $84$ that failed are used in the evaluation. All failures have one or more ground-truth failure causes, in total $163$ causes. The evaluation was performed with $n = 50$ and $\alpha = 0.8$.\footnote{The data used for evaluating the diagnosis algorithm are available at \url{https://zenodo.org/record/4603348}}

    The second experiment follows from the first, namely the diagnosed failures are used for correcting the failed execution samples, which results in a set of synthetic samples. For scoring the corrections, a success prediction model is learned using the data from the first experiment. The synthetic samples, along with the failed executions from which they originated, are then used to create a new success prediction model purely from failed and synthetic experiences; here, the predicted success likelihoods of the failed and synthetic samples are taken to be $0$ and $1$ respectively. We then evaluate this model by repeating the handle grasping action $60$ times, such that we compare two variations of the correction algorithm with different values of the shape parameter $\kappa$, which affects the size of the corrective updates.\footnote{Our implementation of the diagnosis and experience correction algorithms can be found at \url{https://github.com/alex-mitrevski/explainable-robot-execution-models}}

    \subsection{Failure Diagnosis}

    \begin{figure*}[tp]
        \begin{subfigure}[t]{0.31\linewidth}
            \centering
            \includegraphics[width=\linewidth]{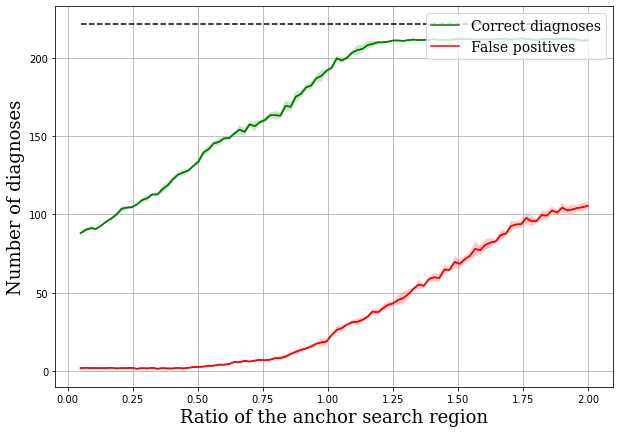}
            \caption{Diagnoses as a function of the ratio of the anchor search region.}
            \label{fig:diagnoses_vs_search_range}
        \end{subfigure}
        \hspace{0.025\textwidth}
        \begin{subfigure}[t]{0.31\linewidth}
            \centering
            \includegraphics[width=\linewidth]{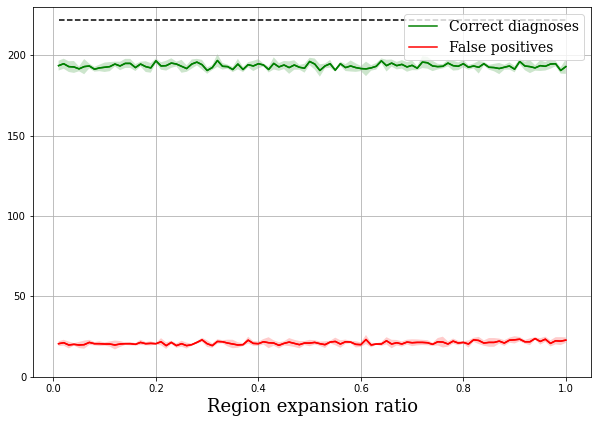}
            \caption{Diagnoses as a function of the region expansion ratio.}
            \label{fig:diagnoses_vs_region_expansion_ratio}
        \end{subfigure}
        \hspace{0.025\textwidth}
        \begin{subfigure}[t]{0.31\linewidth}
            \centering
            \includegraphics[width=\linewidth]{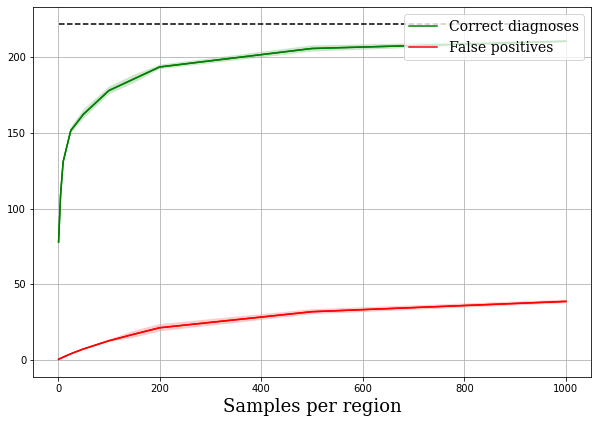}
            \caption{Diagnoses as a function of the number of samples per search region.}
            \label{fig:diagnoses_vs_samples_per_region}
        \end{subfigure}
        \caption{Parameter sensitivity of the diagnosis algorithm. The black dashed line is the total number of ground-truth diagnoses. All results are averaged over five runs, and the shaded regions represent one standard deviation from the mean.}
    \end{figure*}

    We evaluate the diagnosis algorithm by investigating the effect of the algorithm's parameters - the initial $\Sigma$, the region expansion ratio $r$, and the number of samples per region $k_{\max}$ - on the number of correctly identified diagnoses and false positive diagnoses. The evaluation is performed over all $84$ failed executions. A distribution of causes of the failed executions is provided in Table \ref{tab:failure_cause_distribution}.

    \begin{table}[htp]
        \centering
        \caption{Distribution of failure causes}
        \label{tab:failure_cause_distribution}
        \begin{tabular}{p{0.75\linewidth} | >{\raggedleft\arraybackslash}p{0.15\linewidth}}
            \cellcolor{gray!10!white}  \textbf{Failure cause} & \cellcolor{gray!10!white} \textbf{Occurrences} \\\hline
            Too far from the handle & 69 \\\hline
            Collision with drawer   & 7 \\\hline
            Too much to the left    & 16 \\\hline
            Too much to the right   & 19 \\\hline
            Too high                & 19 \\\hline
            Too low                 & 33 \\\hline
            \textbf{Total}          & 163 \\\hline
        \end{tabular}
    \end{table}
    We note that these causes are described by $222$ relations in the ground-truth failure annotations. All results in this section are shown with respect to the total number of relations.

    We first investigate the effect of changing the initial $\Sigma$, namely how the size of the initial search region affects the diagnoses found by the algorithm. For this, we use an anchor set of standard deviations, which is set to $10\%$ of the average size of the handle's bounding box over all executions, and run the algorithm with $100$ ratios of the anchor in the range $[0.05, 2.0]$, keeping $r$ fixed to $0.05$ and $k_{\max}$ fixed to $200$. The results of this evaluation are shown in Fig. \ref{fig:diagnoses_vs_search_range}. As can be seen, the size of the initial $\Sigma$ has a significant effect on the identified diagnoses and, additionally, the number of correct diagnoses is unavoidably accompanied by a significant number of false positive diagnoses. For the handle grasping use case, the algorithm is able to correctly identify most of the failures when the anchor search region is used, at the expense of around $20$ false positive diagnoses as well. As can be seen in the figure, the number of false positives increases with the size of the search region, which is unsurprising since using a larger search region reduces the search granularity.

    Using the anchor search region to initialise $\Sigma$, we next investigate how the region expansion ratio $r$ affects the number of identified diagnoses, such that we run the algorithm with $100$ different expansion ratios in the range $[0.01, 1.0]$, again keeping $k_{\max}$ fixed to $200$. As can be seen in Fig. \ref{fig:diagnoses_vs_region_expansion_ratio}, for a fixed initial $\Sigma$, the expansion ratio does not seem to affect the number of diagnoses, which remains mostly constant over the runs. The effect of this parameter may be more visible for smaller initial values of $\Sigma$, for instance if the initial values in $\Sigma$ are very small and a large value of $r$ is used.

    Finally, using the same initial $\Sigma$ and $r = 0.05$, we investigate how the number of found diagnoses changes with the number of samples $k_{\max}$. Here, we run the algorithm with $10$ different values for $k_{\max}$ in the range $[1, 1000]$. The results are shown in Fig. \ref{fig:diagnoses_vs_samples_per_region}, where it can be seen that small values of $k_{\max}$ generally lead to worse diagnosis results, while larger values allow identifying the failure causes. The effect of larger values saturates; in the handle grasping case, a value of $k_{\max} > 400$ does not change the number of correct diagnoses, but increases the number of false positives.

    The runtime of the algorithm is affected by both $\Sigma$ and $k_{\max}$ and varies depending on the failures to be diagnosed. For $\Sigma$ set to the anchor search region, $r = 0.05$, and $k_{\max} = 200$, it takes about $100s$ to process all $84$ failures, which is a reasonable runtime of about $1s$ per failure.\footnote{This runtime is valid for a Python 3 implementation of the algorithm, running on a machine with an Intel i7 CPU at 2.60GHz and 16GB of RAM.}

    \subsection{Experience Correction}

    We demonstrate the utility of learning from failed executions by evaluating the performance of our robot on the handle grasping action using a success prediction model created from a set of corrected experiences along with the failed samples from which they were extracted. The parameters of the $14$ successful executions are thus only used by the correction algorithm, but are discarded for the evaluation presented in this section. For diagnosis, $\Sigma$ is set to the anchor search region, $r = 0.05$, and $k_{\max} = 200$; for the correction algorithm, $s_{\max} = 10$ is used. It should be noted that, due to its local perturbation nature, the correction algorithm can only correct a subset of the failed experiences,\footnote{Particularly the samples that already violate the preconditions before local perturbations require more than just a local correction.} so only corrections for which the algorithm could find a valid non-zero update are used for creating a success prediction model.

    To investigate how the magnitude of the parameter correction affects the execution success, we generate corrections with two different values of the shape parameter $\kappa$: a value of $\kappa = 2$, which favors parameter values that are directly opposite the falsifying update, and $\kappa = 4$, which favors larger parameter updates. Both corrections sets are used for learning a separate success prediction model, each of which is evaluated over $60$ runs of the handle grasping action. The numbers of corrected experiences used for learning the models and the robot's successes are shown in Table \ref{tab:successful_handle_grasps}.
    \begin{table}[htp]
        \centering
        \caption{Successful handle grasps (out of 60)}
        \label{tab:successful_handle_grasps}
        \begin{tabular}{>{\centering\arraybackslash}p{0.05\linewidth} | >{\centering\arraybackslash}p{0.4\linewidth} | >{\centering\arraybackslash}p{0.4\linewidth}}
            \cellcolor{gray!10!white} $\kappa$ & \cellcolor{gray!10!white} \textbf{Corrected executions} & \cellcolor{gray!10!white} \textbf{Successful grasps} \\\hline
            2 & 28 & 37 \\\hline
            4 & 27 & 22 \\\hline
        \end{tabular}
    \end{table}

    As these results show, even though the models were only learned from corrected executions, the success rate is reasonably high, namely the robot succeeded in about half of the executions over the two experiments. The success rate is higher for $\kappa = 2$ (around $60\%$), which suggests that local corrections are more reliable than larger parameter corrections, which may lead to new, unforeseen failures.


\section{DISCUSSION AND CONCLUSIONS}
\label{sec:discussion}

    In this paper, we presented two sampling-based algorithms that allow a robot to diagnose failed action executions and then correct the parameters of those failed executions, respectively, such that the ultimate objective is to allow robots to learn from failed experiences effectively. Diagnoses are found by perturbing the parameters of a failed execution until the relations of a parameterised execution model are falsified; the identified diagnoses are subsequently used to find alternative action parameterisations that move the parameters away from the failure region. In a handle grasping experiment with a Toyota HSR, we demonstrated that the diagnosis algorithm is able to identify most of the failure causes in a set of failed executions, but the outcomes are affected by the size of the search region and the number of samples used while searching for relation violations. We additionally demonstrated that the robot can learn from only failed and corrected experiences to perform the grasping action with moderate success, but the success rate is affected by the magnitude of the correction.

    The results presented in this paper open up various potential avenues for future work. The purpose of the experience correction procedure is to produce synthetic experiences that can be used for updating the existing success prediction model of a failed action, namely for lifelong model learning. In this paper, we did not address the problem of incorporating such experiences into an existing model, but envision that this process would involve (i) relabelling the failed executions with a low predicted success likelihood, (ii) assigning a high predicted success to the synthetic experiences, and (iii) recreating the model to include both old and synthetic experiences. Before a robot starts applying the updated model, the model should ideally be verified, either in a simulated environment or by a human operator. The failures of interest in this paper were of spatial nature, but it would also be interesting to investigate how the method generalises to failures of other types, such as failures due to incorrectly applied forces, for instance while grasping deformable objects, or temporal failures, such as when pulling objects to a predefined destination. As demonstrated by the experiments, the size of the search region of the diagnosis algorithm affects the quality of the identified diagnoses, which suggests that it may also be worthwhile to consider using procedures for optimising the search region automatically instead of using a heuristic value as we have done in this paper. Finally, since the relations from which an execution model is learned have a direct effect on the diagnosability of failures, the assumption about the mapping between relations and action parameters should be relaxed and the use of a symbol learning method, such as \cite{ames2018}, should be investigated (together with a procedure that translates learned symbols into a form understandable for operators), as this would reduce the burden on designers of execution models.


\addtolength{\textheight}{-12cm}   


\section*{ACKNOWLEDGMENT}

    We thank Santosh Thoduka and Ahmed Abdelrahman for their comments on an earlier draft of this paper.


\bibliographystyle{IEEEtran}
\bibliography{references}

\end{document}